# An Overview of Open-Ended Evolution: Editorial Introduction to the Open-Ended Evolution II Special Issue

Norman Packard*
Protolife, Inc.
  norman.packard@protolife.com

Mark A. Bedau
Reed College

Alastair Channon
Keele University

Takashi Ikegami
University of Tokyo

Steen Rasmussen
University of Southern Denmark
Santa Fe Institute

Kenneth O. Stanley
Uber AI Labs

Tim Taylor
Monash University

**Abstract**  Nature's spectacular inventiveness, reflected in the enormous diversity of form and function displayed by the biosphere, is a feature of life that distinguishes *living* most strongly from *nonliving*. It is, therefore, not surprising that this aspect of life should become a central focus of artificial life. We have known since Darwin that the diversity is produced dynamically, through the process of evolution; this has led life's creative productivity to be called *Open-Ended Evolution* (OEE) in the field. This article introduces the second of two special issues on current research in OEE and provides an overview of the contents of both special issues. Most of the work was presented at a workshop on open-ended evolution that was held as a part of the 2018 Conference on Artificial Life in Tokyo, and much of it had antecedents in two previous workshops on open-ended evolution at artificial life conferences in Cancun and York. We present a simplified categorization of OEE and summarize progress in the field as represented by the articles in this special issue.

## 1  Introduction

The variety of organisms produced by biological evolution is staggering. Recent estimates based on scaling arguments estimate the number of microbial species alone to be $\sim 10^{12}$ [23]. In a very real sense, the entire biosphere has been produced by (or one might say "invented by" [33]) the process of biological evolution. This creative productivity is one of the most striking features of life, and so it has not surprisingly become a focus of artificial life research. Decades of research have explored evolutionary algorithms of various sorts, but so far a full understanding of the creative productivity of evolution (biological or non-biological) has remained out of reach, with regard to both understanding how it has worked to produce the biosphere, and understanding what principles might be used or instantiated outside of biology (the purview of artificial life) to achieve comparable levels of creative productivity.

---

* Corresponding author.

Nature's creative productivity is generally assumed to be ongoing, not only at work over the past 4 billion years, but also at work presently in the biosphere. As far as we know, the same physical, chemical, and biological mechanisms remain at work now as they have in the past, leading most scientists to believe that biological evolution is *open-ended*. This has historically led us to call the study of the creative production of ongoing novelty *open-ended evolution*. As this field of study has developed, it has become clear that processes having the property of ongoing creative productivity are not necessarily biological, chemical, or even evolutionary (at least in a Darwinian sense). This realization leads us to generalize the description of the endeavor to include *open-endedness* in any form.

Not surprisingly, research on open-ended evolution in artificial life has echoes in the biological literature of *evolutionary innovation*. A good overview of recent work on this topic is provided by Hochberg et al. in the Introduction to a theme issue of the *Philosophical Transactions of the Royal Society* on "Process and pattern in innovations from cells to societies" [19]. Hochberg and colleagues state that their ultimate goal is to "understand the central features that drive innovation in all living systems [encompassing biology, culture, and technology], paving the way for a general theory" [19, p. 13]. They define innovations as a subset of all novelties in a system that meet the necessary conditions of (1) representing a qualitative change in a phenotypic trait and (2) having positive fitness. We can immediately see how such a definition might be relevant to the problem of distinguishing "interesting" novelty, a topic we will discuss below. Hochberg et al. go on to distinguish between *performance innovations* that lead to efficiency improvements but do not fundamentally change an organism's ecological niche, and *niche innovations* involving the utilization of a new niche and creating opportunities for adaptive radiation [19, p. 3]. Discussing the importance of the topic of innovation, Hochberg et al. list several challenging questions that remain unanswered, including "is innovation open-ended?" [19, p. 8]. The experience from OEE research suggests that the answer to this question is "not necessarily": The challenge for OEE researchers is to understand the conditions under which it exists, and the mechanisms necessary for achieving it. Moving forward, there is clearly rich potential for a profitable two-way exchange of ideas between those studying evolutionary innovations and those studying OEE.

Open-ended evolution comes in different kinds, and this overview will employ the *Tokyo categories of OEE* (see below). The Tokyo categories are a simplified and modified revision of the categories produced by the York workshop on OEE [38]. We offer and use the Tokyo categories in a provisional and pragmatic spirit. Even if they fix some problems with the York categories, the Tokyo categories still have weaknesses. The authors of this overview reached no consensus about the best way to categorize OEE, and we expect the Tokyo categories to be superseded in the future, even if they are useful enough for today.

A categorization of open-ended evolution provides a big picture of the whole phenomenon, and it allows particular instances to be understood in their relation to other instances. Furthermore, seeing the big picture may help us to spot gaps and guide research to underdeveloped topics. Even provisional and ad hoc organizational tools like the Tokyo categories can provide these benefits.

## 2 Categories for OEE

An important landmark in the attempt to categorize open-ended evolution was the report from a workshop on OEE at York [38]. One of the main conclusions from York was that open-endedness is a variegated concept; there is not a simple single test for the phenomenon, but instead there are different kinds of open-ended evolution, and more than one kind can be exhibited by a single evolving system. The York report summarized the different kinds of OEE as follows:

1. Ongoing generation of adaptive novelty:

   (a) Ongoing generation of new adaptations

   (b) Ongoing generation of new kinds of entities

    (c)  Emergence of major transitions

    (d)  Evolution of evolvability

2.  Ongoing growth of complexity:

    (a)  Ongoing growth of entity complexity

    (b)  Ongoing growth of interaction complexity

We call these the *York categories* of OEE.

The York categories are far from perfect. For example, a new kind of entity might come into existence because it has a new kind of adaptation, so ongoing generation of new kinds of entities (York type 1(b) OEE) could be the same thing as ongoing generation of new adaptations (York type 1(a) OEE). Furthermore, the new kinds of entities (with new adaptations) might also be getting more complex, so York types 1(a) and 1(b) could also be the same thing as York type 2(a) OEE, and similar considerations apply to York type 2(b) OEE.

When we set out to discuss the work in these special issues, these difficulties motivated us to create a simpler and we hope more useful way to categorize OEE. Our revised categories are offered in the same pragmatic spirit as the York list. We sought an organizing framework that helps illuminate the progress in OEE research, knowing full well that this very progress might lead in the future to further category revisions. Our revised categories of OEE consist of the *ongoing generation* of the following four kinds:

1.  Interesting new kinds of entities and interactions

2.  Evolution of evolvability

3.  Major transitions

4.  Semantic evolution

We call these the *Tokyo categories* of OEE because they crystallized when describing the work presented at the OEE workshop in Tokyo.

Figure 1 shows the four kinds of ongoing generation listed in the Tokyo categories (on the left), and it shows the articles in the special issue (on the right). The lines indicate which articles concern which Tokyo categories of OEE. The following section briefly describes each Tokyo category and then discusses those articles that concern it.

The Tokyo categories contain one brand-new category—*semantic evolution* (see below)—but the other Tokyo categories appear in some form on the York list. For example, Tokyo category type 1 OEE covers what were separated on the York list as the ongoing generation of new kinds of entities, new adaptations, and increasingly complex entities. In the end the Tokyo categories are simpler in that the York hierarchical structure is flattened and some York categories have been merged. Like the York categories, the Tokyo categories do not partition examples of OEE into mutually exclusive sets, so the very same evolutionary process could exemplify more than one Tokyo category. This explains why Figure 1 links some articles to more than one Tokyo category. The Tokyo list calls attention to four kinds of OEE that seem especially important and useful if we judge by the work collected in this special issue.

The term "interesting" in the description of type 1 OEE (new kinds of entities or interactions) deserves special mention, because it makes type 1 OEE vague and subjective. A more precise and objective substitute would improve the description of this category of OEE. We will see below that some work in the special issue shows why such a qualification is needed and how it can be made more precise and objective.

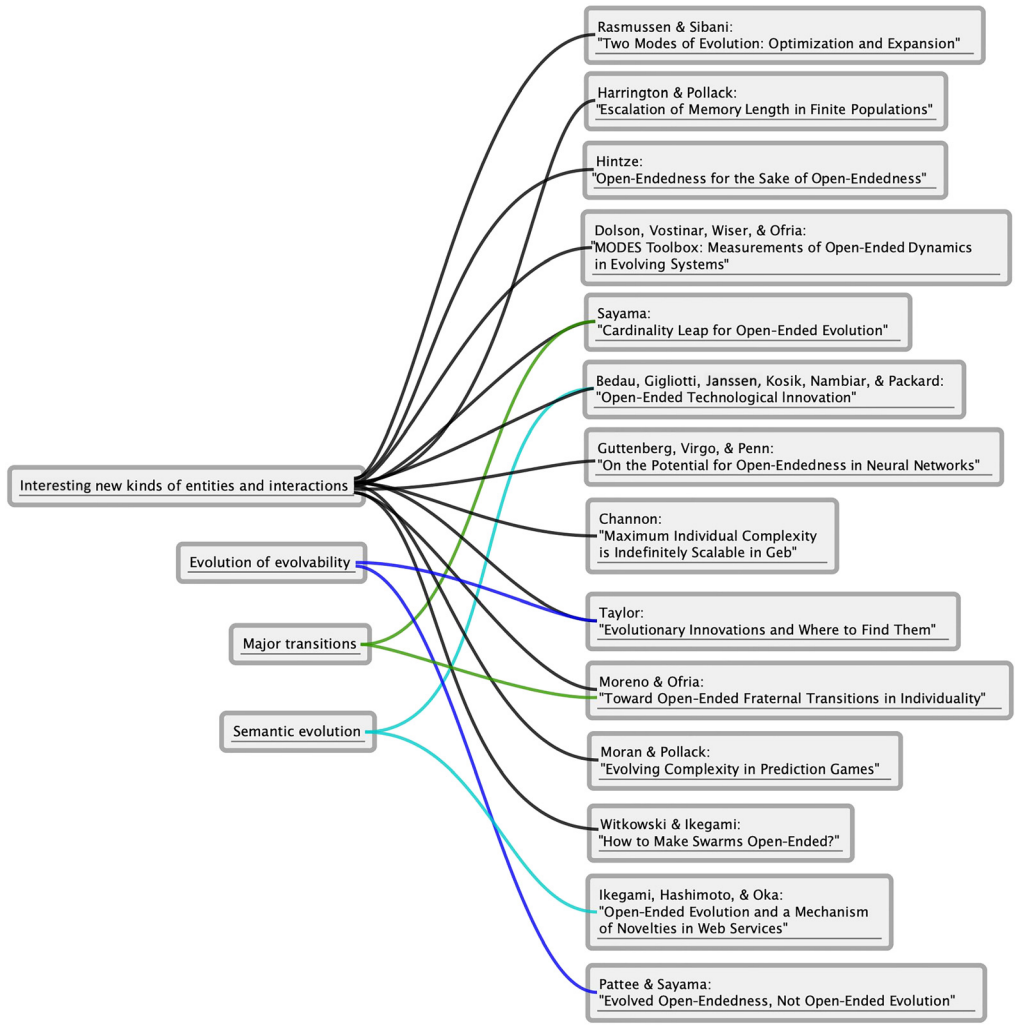

Figure 1. The Tokyo categories, with links to articles in the special issues, according to category relevance. Note that some articles are linked to more than one category.

## 3  Progress in OEE Research

The *Artificial Life* journal focuses on OEE in two special issues, 25(1) and the current issue 25(2). In this section we review all the contributions to these special issues on OEE, and we discuss progress in our understanding of each of the four Tokyo categories. We also discuss progress on certain overarching issues that concern more than one kind of OEE.

### 3.1  Interesting New Entities and New Interactions

Generation of new entities is often seen in various ALife models, so it is no surprise that several articles address this aspect of open-endedness. As seen in Figure 1, models with new entities are often accompanied by new interactions.

Both Harrington and Pollack's contribution [13] and Moran and Pollack's contribution [25] model different versions of an evolving population of entities competing with each other through game play. The entities evolve by increasing entity complexity by evolving each entity's game-playing strategy. Both models provide an elegant version of OEE via increase in entity complexity.

Although ALife research has focused for a long time on Tokyo type 1 OEE (ongoing generation of interesting new kinds of entities and interactions), quantification has been elusive because of the vague and subjective term "interesting." This difficulty is considered in a formal setting by Banzhaf et al. [4]; additionally, Stanley and Soros [34] explicitly point out that interestingness is subjective. An important open question about OEE is how to replace this vague and subjective term, and two contributions to these special issues have provided new illumination.

The contribution by Hintze presents a tantalizingly simple system that produces ongoing generation of new entities [18]. The resulting diversity is trivial, in the sense that even though the entities produced are different, they are statistically very much the same. The "newness" of the entities is neither geometric, nor dynamic, nor functional. This model shows why we need a characterization of which new kinds of entities are so "interesting" that their ongoing generation counts as open-ended evolution. Hintze concludes that sharper definitions of complexity of entities and interactions are necessary to proceed beyond his explicit version of simple OEE.

Bedau et al. present an empirical analysis of technological evolution, where ongoing production of new industry clusters is observed [5]. The quality of being interesting has a concrete realization in this work: New structure is recognized as interesting if it is statistically significant enough to be detected by a feature-detecting learning algorithm that marches in parallel with the data production process provided by the stream of human-produced patents, itself an open-ended process. This provides a definite, measurable quantification of interestingness, at least for their OEE context.

A key feature of some kinds of biological open-endedness is that entities interact with each other on one scale and then form new entities on a different (typically larger) scale. One version of this phenomenon was elaborated with the notion of *dynamical hierarchies* [30]. The term "dynamical hierarchy" was coined to describe emergence of a hierarchy of physical and functional structure during a dynamical process, in the context of large-scale, complex molecular dynamics simulations: atomic entities joining to form molecules, molecules joining to form supramolecular structures, and so on. Biological systems display emergence of even more elaborate hierarchies, with molecular entities forming cells, cells forming organisms, organisms forming ecologies, and so on. Emergence of rich hierarchical structure is one concrete example of interesting new entities and interactions, and is a "holy grail" for ALife models.

These special issues contain significant progress on this front. One example is Sayama's "cardinality leap" for entities observed in an artificial chemistry model termed *Hash Chemistry* [32]. Hash Chemistry has the built-in limitation of a preordained fitness function (Mathemeetica's `Hash` function), which makes it a model with an extrinsic fitness definition [27]. The extrinsic fitness might be considered a shortcoming of the model, but the interesting feature of the `Hash` function is that it can provide fitness values not only for individual entities in the model, but also for aggregates of entities (spatially nearby). The ability of the `Hash` function to evaluate aggregates is a mechanism for generating open-endedness, reflected in a "cardinality leap" of the possibilities available to the evolutionary process. The emergence of fit aggregates in Hash Chemistry is a form of evolutionary emergence of hierarchy.

Another interesting example of emerging hierarchical structure is provided by Moreno and Ofria's DISHTINY (distributed hierarchical transitions in individuality) platform [26]. The structure of the model naturally provides for evolutionary exploration of both upward and downward hierarchical structure. An example of the structure for evolved signaling networks is shown in Figure 2.

Emergence of large-scale spatial structure is another especially interesting kind of open-ended evolution. Here we consider it as an example of Tokyo category 1 (interesting new kinds of entities and interactions), but it may also deserve separate consideration. A model that displays the emergence of rich large-scale spatial structure is the evolving *boids* of Witkowski and Ikegami [39]. Not only do the boids' interactions evolve, but the large-scale structure of the flock also evolves.

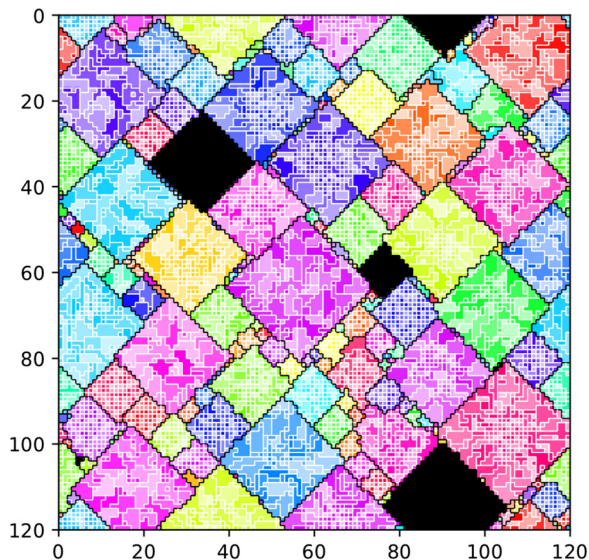

Figure 2. Evolved signaling networks, from the DISHTINY platform [26] (figure used with permission). Note emergence of a variety of cooperative structures on different length scales. See [26] for an explanation of the figure.

Community formation is a specific form of emergent interaction complexity. This is one aspect of properties that emerge in an empirical study of web services, presented in the contribution of Ikegami, Hashimoto, and Oka [20].

Ackley has argued that indefinitely scalable complexity is one requirement for certain kinds of OEE [1], and Moreno and Ofria consider the issue of indefinitely scalable complexity in their DISHTINY model (see Figure 2). Channon's contribution to this special issue [8] succinctly describes how indefinite scalability is linked to accumulation of adaptive success. Essentially, indefinitely scalable complexity implies that the "complexity carrying capacity" of a chunk of space and time filled with interacting components should, for systems that display OEE, be a thermodynamically extensive quantity, which is to say it will scale with the space-time volume of the world. Indefinitely scalable complexity has become a hallmark of OEE systems, essentially considered a necessary condition. More specifically, Channon presents evidence for indefinite scalability of maximum individual complexity in Geb [6, 7] (based on Harvey's SAGA framework for incremental artificial evolution [14–17] but with coevolutionary feedback arising via biotic selection rather than abiotic fitness functions) when scaling both world length (which bounds population size) and the maximum number of neurons per individual together. Further, maximum individual complexity is shown to scale logarithmically with (the lower of) these parameters, raising the general question of how to improve on this order of complexity growth.

## 3.2   Evolution of Evolvability

The primary contribution on evolution of evolvability is provided by Pattee and Sayama [28]. They take a novel perspective, turning open-ended evolution on its head, and considering how open-endedness itself may be a product of evolution. This general discussion echoes more specific studies of evolvability of evolvability, such as evolutionary innovations that may increase the evolvability (capacity for adaptive evolution) of their lineages [29, 35].

Taylor's contribution addresses a form of evolution of evolvability, when he considers the implementation of intrinsically coded evolvable genetic operators. In systems where the evolutionary process is implemented in this way, "not only might the process evolve, but the *evolvability of the process* might itself evolve" [37, p. 216].

### 3.3  Major Transitions

The term "major transitions in evolution" was introduced by Maynard Smith and Szathmáry [24] (and recently revisited by Szathmáry [36]). Major transitions are characterized by an emergent hierarchy, with each new level in the hierarchy consisting of a new population of reproducing and evolving entities—thus a particular kind of dynamical hierarchy [30]. A major transition in evolution is preceded by the evolution of one or several distinct kinds of reproducing entity. Eventually certain groups of those entities come to interact very tightly, and they become members of a new population of higher-level reproducing wholes. Entities in the old lower-level population become parts of the new wholes, but they cannot reproduce independently. Now the process repeats once more. Certain groups in the population of new wholes come to interact very tightly, and they become new, even higher-level wholes that reproduce and form an even higher-level population, and so on. Maynard Smith and Szathmáry hold that the major transitions in evolution they survey are quite contingent; they could easily not have happened, and there may be no more major transitions. In this view, the existence of some major transitions in evolution is not necessarily any kind of OEE. But major transitions can spur many further adaptations and help make evolution open-ended.

An alternate view might accept the contingency of particular major transitions, as described by Maynard Smith and Szathmáry, but regard the presence of major transitions (as envisioned by them) to be inevitable in open-ended evolution. In any case, ongoing production of major transitions would be an especially impressive kind of OEE.

The contributions to these special issues that shed the most light on emergence of major transitions are Sayama's "cardinality leap" [32] and Moreno and Ofria's "open-ended fraternal transitions" [26]. The emergence of hierarchical structure described above for these models opens the possibility that cooperation of tightly linked entities might constitute a version of the tightly interacting groups that constitute a new population at a higher level.

### 3.4  Semantic Evolution

Ikegami, Hashimoto, and Oka [20] suggest a new category of open-ended evolution having to do with the evolution of semantic relationships within an evolving system: *semantic evolution*. An example of this may be seen explicitly in their contribution to this issue, where we see evolving meaning of tags by creating new combinations of the tags in the evolution of web services.

Another version of semantic evolution may be seen in Bedau et al.'s analysis of technological evolution [5], where keywords characterizing industry clusters shift in meaning and importance with time. Perhaps semantic evolution could be seen as the evolution of interactions in a semantic network, and so perhaps semantic evolution is a special case of Tokyo type 1 OEE (ongoing generation of interesting new kinds of entities and interactions). Nevertheless, the evolution of semantic relationships may be sufficiently novel and important to deserve its own category.

Both examples of semantic evolution are non-biological in nature, and indeed they come from empirical analyses of human-produced evolving systems. The clear display of semantic evolution in these systems might raise the question whether semantic evolution exists in other evolving systems unrelated to natural language. For example, does a semantics for biochemistry come from thinking of "meaning" as chemical function? If so, might we observe "semantic" evolution as the chemical function of certain molecules change with their chemical context? These aspects of OEE in biochemical systems have points of connection with the biosemiotics literature [21], from which some productive synergies might arise in future work.

### 3.5  Beyond Categories

Some of the contributions in these issues address very general properties of open-endedness, not limited to specific categories of OEE. The contribution of Taylor explores and categorizes various routes by which the capacity for open-endedness can be introduced into biological and artificial evolutionary systems [37]. And the contribution of Guttenberg, Virgo, and Penn [12] makes a powerful analogy between evolving systems and deep learning systems that are, at their heart, using gradient descent.

Rasmussen and Sibani make a distinction between two different modes of evolution at work in (presumably) any version of open-ended evolution: optimization and expansion [31]. Their distinction resembles the well-known tradeoff between exploitation and exploration in many machine learning systems [2, 3]. Exploitation (of a learned model) is similar to optimization, and exploration is related to expansion, but with an important difference: Exploration in machine learning always involves search in a fixed-sized state space, while evolutionary expansion always involves an expansion of the state space involved.

One problem for the statistical detection of open-endedness in evolving systems is that statistics can be gathered only for the kinds of entities that are specified in advance. But once the kinds of entities are prespecified, how can the production of new and different kinds of entities be recognized? This issue is often approached by defining a very large space of possible entities, and initializing an evolving system to populate only a small piece of the space. This enables open-ended dynamics to be observed for a while, until the limits of the space are exhausted. Formal aspects of this conundrum are discussed explicitly in the contribution of Taylor [37].

The contribution of Bedau et al. provides an alternative pragmatic and empirical approach to the problem of identifying new and unanticipated kinds of entities. They view the evolving system through a lens consisting of an algorithm that learns structure over time, so that the components can be learned dynamically and after the fact [5]. Statistics are collected on the new kinds of entities that have been detected, and those kinds of entities can be temporally related but constantly changing.

A set of tools for analysis of open-ended dynamics is presented by Dolson et al. in their article describing the MODES toolbox [10], used to measure the change potential, novelty potential, complexity potential, and ecological potential of an evolving system. The authors provide detailed algorithms and a C++ implementation of their toolbox. They take a different approach to the statistical conundrum of OEE: Rather than merely gathering statistics, they compute quantities like change potential, novelty potential, and complexity potential as estimates of how much evolutionary "space" is available to the evolving system.

## 4    Conclusions

The first and foremost conclusion from the articles in these special issues is that our work is not done. The degree of open-endedness displayed by biological evolution remains out of reach of today's ALife models, and we don't understand the mechanisms behind OEE well enough to engineer systems that display that degree of open-endedness. Nevertheless, the progress presented in these special issues is notable and marks an increase in our understanding of OEE, how to measure it, and how to build technologies that embody it.

The Tokyo categories of OEE we used here are only provisional; we may expect them to change as we understand more. The biological literature discussed above [19] may reveal lacunae in our categories, for it distinguishes *performance* and *niche* innovations. Into which category of OEE should we place niche formation and niche innovation? Perhaps they could be viewed as a type of growing interaction complexity, or perhaps they should be viewed as akin to the evolutionary "optimization" and "expansion" distinguished by Rasmussen and Sibani [31].

Besides looking for insights into open-endedness in biology with its emphasis on adaptation and selection, it is important to note that open-endedness can also occur in artificial systems that seem free from adaptation and selection. For example, algorithms like novelty search [22] generate ongoing novelty without any evident adaptive pressure.

One might speculate that the analysis of open-ended evolution and open-ended dynamics will also eventually bring into focus three further properties of open-ended systems:

- *Functionality.* Components that become important in open-ended systems typically interact with other components and can be considered to "perform a function." For example, catalytic feedback loops perform a function in chemistry. Does functionality play an important role in certain kinds of OEE? Is it quantifiable and observable? Clearly not every component of an open-ended system need be functional, but perhaps open-ended growth of functionality is another important kind of OEE. Taylor provides some initial discussion of these questions in the final section of his contribution [37].

- *Information processing and computation.* The biological world, and more recently the social–computational world, seem to generate more computational sophistication with time. It certainly seems abundantly clear that there is more information contained, communicated, and processed by the contemporary biosphere than there was in the biosphere four billion years ago. Is some aspect of this increase in information processing a hallmark of certain kinds of OEE? Is it quantifiable and measurable? A positive answer might be found in the application of learning algorithms coupled to data streams produced by open-ended systems; examples include Crutchfield's epsilon-machines [9, 11] and the contribution of Bedau et al. to these special issues [5]. Is increase in information processing and/or computation a necessary condition for some particularly important category of OEE? Channon's linkage of indefinitely scalable complexity [1] with unbounded accumulation of adaptive success hints at an affirmative answer to this question [8].

- *Nonequilibrium thermodynamics.* Thermodynamically, the biosphere is clearly an example of structure produced by a nonequilibrium system, with solar energy flowing onto the earth, making its way through the biosphere until it reaches its lowest entropy state, the heat bath. We would like to relate features of open-ended evolution to this thermodynamic process. Is this possible? Many ALife models for OEE do not have precise physical or chemical analogues. Are there proxies within the models that might enable formulation of "laws" governing certain kinds of OEE that would carry over to real chemical and physical processes? If we want to move toward engineering specific categories of OEE in the world of chemistry and physics, we may need the answers to these questions.

## Acknowledgments

The editors would like to thank the ALife community that participated in the production of these special issues of the *Artificial Life* journal, in roles of both writing contributions and reviewing them.